\documentclass[a4paper,conference]{IEEEtran}
\ifCLASSINFOpdf
\else
\fi
\usepackage{amsmath}

\usepackage{amsmath}
\usepackage{amssymb}
\usepackage{siunitx}
\usepackage{multirow}
\usepackage{color}
\usepackage[nocompress]{cite}
\usepackage{graphicx}

\begin{document}
%
\title{Domain Translation with Conditional GANs: \\ from Depth to RGB Face-to-Face}

\author{\IEEEauthorblockN{Matteo Fabbri \quad Guido Borghi \quad Fabio Lanzi \quad Roberto Vezzani \quad Simone Calderara \quad Rita Cucchiara\\}
\IEEEauthorblockA{\\Department of Engineering ``Enzo Ferrari''\\
University of Modena and Reggio Emilia\\
via Vivarelli 10 Modena 41125, Italy\\
\{name.surname\}@unimore.it}}


%


\maketitle

\begin{abstract}
Can faces acquired by low-cost depth sensors be useful to catch some characteristic details of the face? Typically the answer is no. However, new deep architectures can generate RGB images from data acquired in a different modality, such as depth data. In this paper, we propose a new \textit{Deterministic Conditional GAN}, trained on annotated RGB-D face datasets, effective for a face-to-face translation from depth to RGB. Although the network cannot reconstruct the exact somatic features for unknown individual faces, it is capable to reconstruct plausible faces; their appearance is accurate enough to be used in many pattern recognition tasks. In fact, we test the network capability to hallucinate with some \textit{Perceptual Probes}, as for instance face aspect classification or landmark detection. Depth face can be used in spite of the correspondent RGB images, that often are not available due to difficult luminance conditions. Experimental results are very promising and are as far as better than previously proposed approaches: this domain translation can constitute a new way to exploit depth data in new future applications.
\end{abstract}


%
\IEEEpeerreviewmaketitle

\section{Introduction}
Generative Adversarial Networks (GANs) have been adopted as a viable and efficient solution for the Image-to-Image translation task, or rather the ability to transform images into other images across domains, according to a specific training set.
Initially, Autoencoders, and in particular Convolutional Autoencoders \cite{masci2011stacked}, have been investigated and designed for several image processing tasks, such as image restoration \cite{mao2016image}, deblurring \cite{bigdeli2017image}, and for image transformations such as image inpainting \cite{guillemot2014image} or image style transformation. They have been used also as transfer learning mechanism for the Domain Transfer task: for sensor to image transformation \cite{singhtransforming} or from depth to gray-level images of faces \cite{borghi17cvpr}.
As mentioned above, the Goodfellow's proposal of GANs \cite{goodfellow2014generative} became the reference architecture for unsupervised generative modeling and for sampling new images from the underlying distribution of an unlabeled dataset by exploiting the joint capabilities of a Generative and a Discriminative Networks \cite{radford2015unsupervised}.
Furthermore, Conditional GANs \cite{isola2016image,matteo2017generative} provided conditional generative models by conditioning the sampling process with a partially observed input image. 
Several experiments show the power and effectiveness of conditional GANs, as for instance to improve resolution or to provide de-occlusion of images of people \cite{matteo2017generative}.\\
\begin{figure}[t]
    \centering
    \includegraphics[width=0.95\columnwidth]{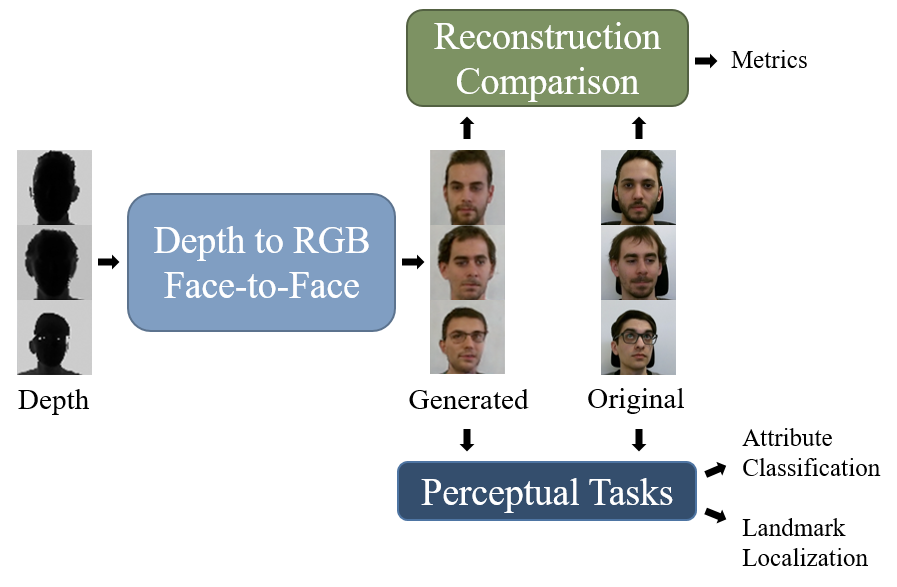}
    \caption{Overview of Reconstruction Comparison and Probe Perceptual Tasks for performance evaluation.}
    \label{fig:probe} 
\end{figure}

\begin{figure*}[th!]
    \centering
    \includegraphics[width=0.75\linewidth]{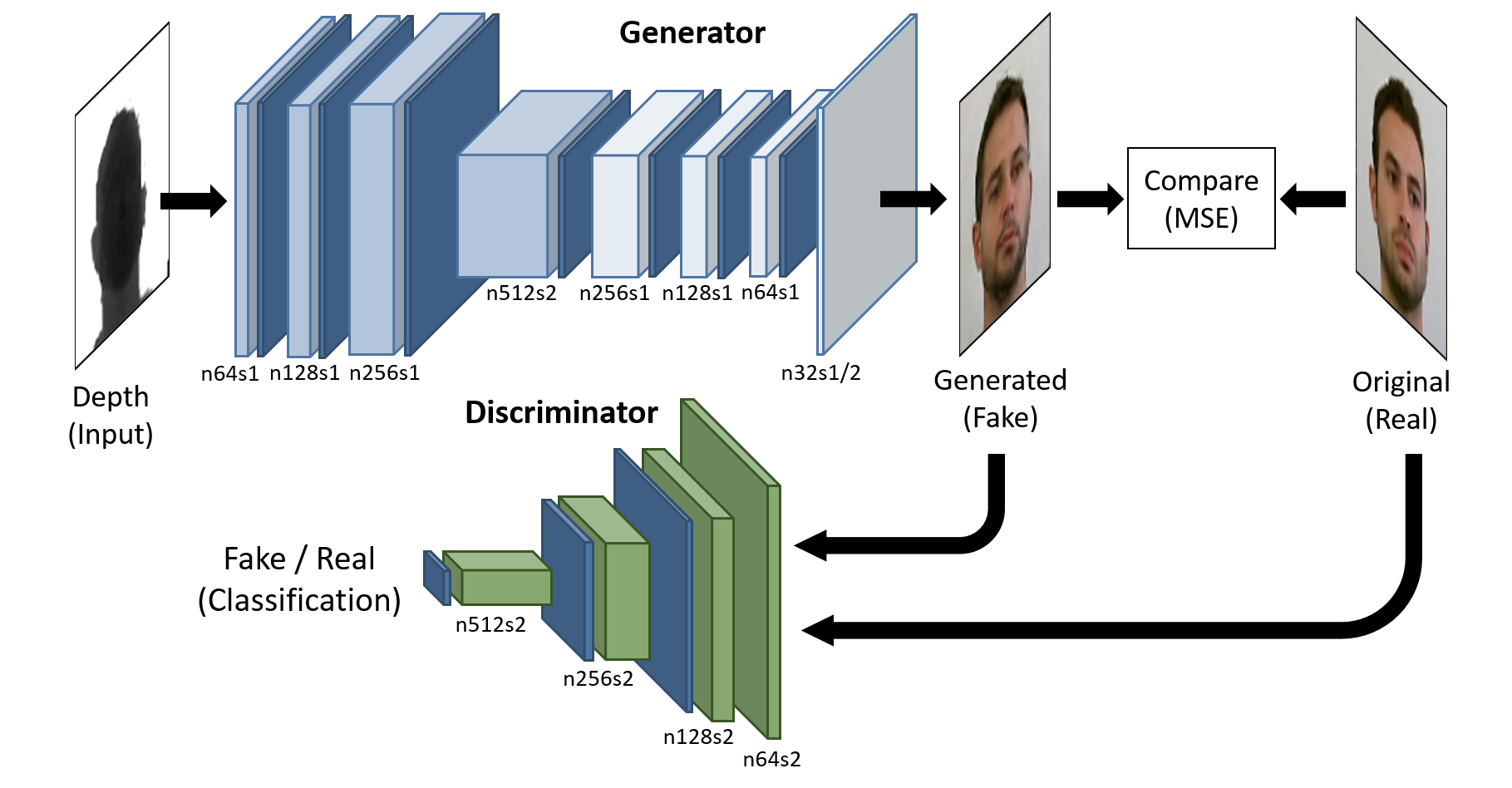}
    \caption{Training schedule for Conditional GANs. The Discriminator learns to classify between generated fake images and real images while the Generator learns to fool the Discriminator. For each layer, the image provide information about number of filters (n) and stride (s).}
    \label{fig:gan} 
\end{figure*}

In this work, we explore the capability of face-to-face domain translation exploiting conditional GANs.
The ability of a network to hallucinate and define a face aspect (in color or gray level), starting from a range map, could be a useful basic step for many computer vision and pattern recognition tasks, from biometric to expression recognition, from head pose estimation to interaction, especially in those contexts where intensity or color images cannot be recorded, for instance when shadows, light variations or darkness make the luminance and color acquisition not feasible enough. 
Our contribution is the definition of a \textit{Conditional Generative Adversarial Network} that, starting from an annotated dataset with coupled depth and RGB faces (acquired by RGB-D sensors), learns to generate a plausible RGB face from solely the depth data. 
The network learns a proper transformation across the color and depth domains. Nevertheless, in generative settings, the result is likely a plausible face which could be qualitatively satisfactory (\textit{e.g.}, the Discriminator network is fooled by it) but it is objectively difficult to properly measure the adherence to the conditioned input. \\
Therefore, another important contribution of our proposal is the adoption of some vision tasks as \textit{Perceptual Probes} for performance evaluation, under the assumption that the domain translation task is viewed as an initial step of a more complex visual recognition task. We assess that the face-to-face translation is acceptable if the new generated RGB face (from depth input) exhibits similar proprieties of other RGB-native faces in the selected probe perceptual tasks (\textit{i.e.}, categorical attributes are maintained across domains). 
In accordance with this assumption, we will provide several experiments to test the proposed solution: we will use two different perceptual probes -- namely, a network for face attribute classification and a method for landmark extraction -- and we will evaluate how these tasks perform on generated faces. The overview of our Probe Perceptual Task is depicted in Figure \ref{fig:probe}.
Results are really encouraging so that this approach could be a first attempt to ``see and recognize faces in the dark'', in analogy to how blind people captures the appearance only by touching a face and sensing the depth shape.

\section{Related works}
\textbf{GAN for Image-to-Image translation.} GANs have been defined very recently and tested in several contexts. Our work is inspired by the first idea of Goodfellow \textit{et al.}, of Generative Adversarial Networks \cite{goodfellow2014generative} with some variant in terms of conditional and discriminative GANs.
GANs have been successfully used for Image-to-Image translation; they have been initially presented in \cite{isola2016image} and then applied to some contexts as unpaired image to image translation \cite{zhu2017unpaired}.
A previous work starting from the same depth images has designed an Autoencoder to create gray-level faces from depth, with the final goal of head estimation \cite{borghi17cvpr}. An extension of this work is performed in \cite{borghi2017face} where a GAN is trained for the same final goal.
In this paper, we compare our architecture with \cite{isola2016image}, using a similar dataset, other datasets, and with some probe perceptual tasks.\\

\begin{figure*}[th!]
    \centering
    \includegraphics[width=0.95\linewidth]{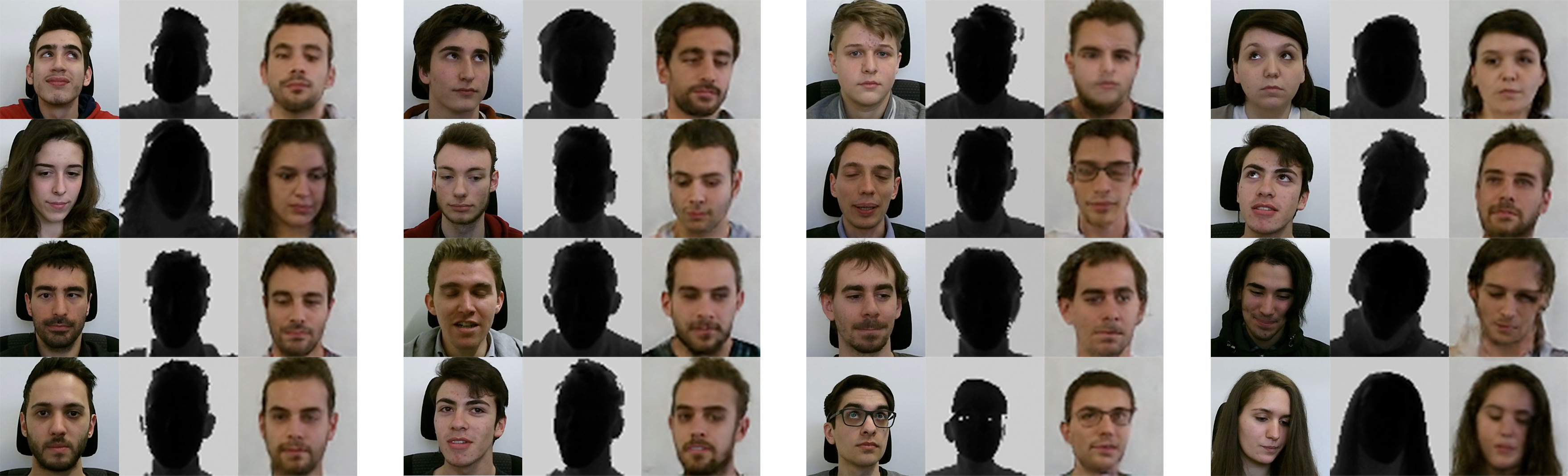}
    \caption{Best results on the \textit{MotorMark} \cite{frigieri2017fast} dataset. For each triplets of images: (Leftmost) the original image; (Middle) the input depth map; (Rightmost) the Generated face image.}
    \label{fig:copertina} 
\end{figure*}

\textbf{Depth Maps and Deep Learning.}
The latest spread of high-quality, cheap and accurate commercial depth sensors has encouraged the researchers of the computer vision community. Depth data are a useful source of information especially for systems that have to work in presence of darkness or dramatic light changes. Besides, recent depth sensors usually exploit infrared lights instead of lasers, so their use is safer for humans.\\ 
In the literature, the potentiality of depth images used as input for deep learning approaches has not been fully investigated yet. Only recently, Convolutional Neural Networks (CNNs) and depth maps have been exploited for various tasks, like head pose estimation \cite{borghi17cvpr,venturelli2017deep}, facial landmark detection \cite{frigieri2017fast}, head detection and obstacle detection \cite{lee2012intelligent}. Various type of deep architecture have been investigated, like LSTM \cite{hochreiter1997long} or Siamese networks \cite{venturellisiamese}. The importance of this source of information is proved by the presence of works that aim to retrieve depth starting from monocular RGB images \cite{saxena2006learning,saxena2009make3d}.   

\section{Proposed Method}
A general view of the proposed method is depicted in Figure \ref{fig:gan}. It consists of a GAN trained and tested on two different datasets, detailed in the following section.\\
GANs are generative models that learn a mapping from random noise vector $z$ to output image $y$: $G:z\rightarrow y$ \cite{goodfellow2014generative}. Conditional GANs instead are generative models introduced by \cite{isola2016image} that learn a mapping from an observed image $x$ and random noise $z$ to an output image $y$: $G:\{x,z\}\rightarrow y$. Like GANs, they are composed of two components: a Generator $G$ and a Discriminator $D$. The Generator $G$ is trained to generate outputs that are indistinguishable from ``real'' by the adversarially trained Discriminator $D$ which is trained to recognize the Generator's ``fake'' images from the ``real'' ones.

Using random noise as input, the generator $G$ creates completely new samples, drawn from a probability distribution that approximates the distribution of the training data. This procedure leads to a non-deterministic behavior, that is undesired for our goal. By removing the noise $z$, the probability distribution approximated by the model becomes a delta function with the property of preserving a deterministic behavior. Deterministic Conditional GANs (det-cGAN) thus learn a mapping from observed image $x$ to output image $y$: $G:x\rightarrow y$.

\subsection{Framework}
The main goal is to train a generative function $G$ capable of estimating the RGB face appearance $I^{gen}$ from the corresponding depth input map $I^{dpt}$ with the objective of reproducing the original image $I^{rgb}$ associated with the depth map. To this aim, we train a Generator Network as a feed-forward CNN $G_{\theta_{g}}$ with parameters $\theta_{g}$. For N training pairs images $(I^{dpt},I^{rgb})$ we solve:
\begin{equation}
\label{eq:a}
	\hat{\theta}_g = \arg \min_{{\theta}_g} \frac{1}{N} \sum_{n=1}^N Loss_{G} \left( G_{{\theta}_g} (I^{dpt}_n) , I^{rgb}_n \right).
\end{equation}
We obtained $\hat{\theta}_g$ by minimizing the loss function defined at the end of this subsection.
Following the det-cGAN paradigm we further define a Discriminator Network $D_{\theta_{d}}$ with parameters $\theta_{d}$ that we train alongside $G_{\theta_{g}}$ with the aim of solving the adversarial min-max problem:
\begin{multline}
\label{eq:b}
	\min_{\theta_{g}} \max_{\theta_{d}} \mathbb{E}_{I^{rgb} \sim p_{data}(I^{rgb})}[\log{D(I^{rgb})}] \\
	+ \mathbb{E}_{I^{gen} \sim p_{gen}(I^{gen})}[\log{1 - D(G(I^{dpt}))}]
\end{multline}
\noindent where $D(I^{rgb})$ is the probability of $I^{rgb}$ being a ``real'' image while $1 - D(G(I^{dpt}))$ is the probability of $G(I^{dpt})$ being a ``fake'' image. 
The main idea behind this min-max formulation is that it gives the possibility to train a generative model $G$ with the target of fooling the discriminator $D$, which is adversarially trained to distinguish between generated ``fake'' images and ``real'' ones. With this approach, we achieve a generative model capable of learning solutions that are highly similar to ``real'' images, thus indistinguishable from the Discriminator $D$.

\begin{table*}[t]
\centering
\caption{Evaluation metrics computed on the generated RGB face images with \textit{MotorMark} dataset. Starting from left are reported $L1$ and $L2$ distances, absolute and squared error differences, root-mean-squared error and finally the percentage of pixels under a defined threshold. Further details are reported in \cite{NIPS2014}}
\small
\begin{tabular}{c||cc|cc|ccc|ccc}
\hline
\multirow{2}{*}{\textbf{Method}} &\multicolumn{2}{c|}{\textbf{Norm} $\downarrow$} &\multicolumn{2}{c|}{\textbf{Difference} $\downarrow$}  &\multicolumn{3}{c|}{\textbf{RMSE} $\downarrow$} &\multicolumn{3}{c}{\textbf{Threshold} $\uparrow$} \\
\cline{2-11}
 &$L_1$ &$L_2$ &Abs  &{\footnotesize Squared}  &linear &log &scale-inv &1.25 &2.5 &3.75 \\ \hline
\hline
Autoencoder &39.80 &6327 &2.21 &273.33 &58.74 &1.248 &1.791 &1.389 &1.878 &2.120 \\ \cline{1-1} \cline{2-11}
pix2pix \cite{isola2016image} & 37.77 & 6150 & 2.06 & 253.11 & 56.01 & 1.240 & 1.846 & 1.400 & 1.882 & 2.157\\ \cline{1-1} \cline{2-11}
\textbf{Our}&\textbf{37.12} &\textbf{6021} &\textbf{2.05} &\textbf{245.88} &\textbf{54.86} &\textbf{1.222} &\textbf{1.749} &\textbf{1.423} &\textbf{1.914} &\textbf{2.188} \\ \hline
\hline
Our (Binary Maps)	& 43.58 & 6868 & 2.45 & 320.93 & 62.53 & 1.320 & 1.830 & 1.319 & 1.778 & 2.047 \\ \cline{1-1} \cline{2-11}
\end{tabular}
\label{tab:metrics}
\end{table*}

As a possible drawback, those solutions could be highly realistic thanks to $D$ but unrelated to the input. A generated output could be a realistic face image with very different visual attributes and different pose with respect to the original image. This setup does not guarantee, for example, that a depth map of a girl with wavy hair looking to the right will generate an RGB image preserving those features. In order to tackle this problem, we mixed the Generator loss function $Loss_{G}$ with a more canonical loss such as MSE. Borrowing the idea from \cite{matteo2017generative}, we propose a Generator loss that is a weighted combination of two components:
\begin{equation}
\label{eq:c}
	Loss_{G} = \lambda Loss_{MSE} + Loss_{adv}
\end{equation}
where $Loss_{MSE}$ is calculated using the mean squared errors of prediction (MSE) which measure the discrepancy between the generated image $I^{gen}$ and the ground truth image $I^{rgb}$ associated with the corresponding input depth map $I^{dpt}$. The MSE component is subject to a multiplication factor $\lambda$ which controls its impact during training.
The $Loss_{adv}$ component is the actual adversarial loss of the framework which encourages the Generator to produce perceptually good solutions that reside in the manifold of face images. The loss is defined as follows:
\begin{equation}
\label{eq:d}
	Loss_{adv} = \sum_{n=1}^N -\log(D(G(I^{dpt})))
\end{equation}
where $D(G(I^{dpt})$ is the probability of the Discriminator labeling the generated image $G(I^{dpt})$ as being a ``real'' image. Rather than training the Generator to minimize $\log(1-D(G(I^{dpt})))$ we train $G$ to minimize $\log(D(G(I^{dpt})))$. This objective provides strongest gradients early in training \cite{goodfellow2014generative}.
The combination of those two component grants the required behavior: the Generator has not only to fool the Discriminator but has to be near the ground truth output in an MSE sense.

\subsection{Architecture}

The task of Image-to-Image translation can be expressed as finding a mapping between two images. In particular, for the specific problem we are considering, the two images share the same underlying structure despite differing in surface appearance. Therefore, the structure in the input depth image is roughly aligned with the structure in the output RGB image. In fact, both images are representing the same subject in the same pose thus details like mouth, eyes, and nose share the same location through the two images. The generator architecture was designed following those considerations.

A recent solution \cite{isola2016image} to this task adopted the ``U-Net'' \cite{ronneberger2015u} architecture with skip connections between mirrored layers in the encoder and decoder segments in order to shuttle low-level information between input and output directly across the network. We found this solution less profitable because the strictly underlying structural coherence between input and output makes the network use the skip connections to jump at easier but not optimal solutions and ignoring the main network flow.

Consequently, our architecture implementation follows the \textit{FfD} implementation in \cite{borghi2017face}. We relaxed the structure of the classical hourglass architecture performing less upsampling and downsampling operations in order to preserve the structural coherence between input and output. We found that using the half of feature maps described in \cite{borghi2017face} at each layer in both Generator and Discriminator networks sped up the training without a significant reduction of qualitative performance.

We propose the Generator's architecture depicted in Figure \ref{fig:gan}. Specifically, in the encoder, we used three convolutions followed by a strided convolution (with stride 2, in order to reduce the image resolution). The decoder uses three convolutions followed by a fractionally strided convolution (also known in literature as transposed convolutions) with stride $1/2$ to increase the resolution, and a final convolution. Leaky ReLU is adopted as activation function in the encoding stack while ReLU is preferred in the decoding stack. Batch normalization layers are adopted before each activation, except for the last convolutional layer which uses the Tanh activation. The number of filters follows a power of 2 pattern: from 64 to 512 in the encoder and from 256 to 32 in the decoder. All convolutions use a kernel of size 5 $\times$ 5.
The Discriminator architecture is similar to the Generator's encoder in terms of number of filters and activations functions but uses only strided convolutional layers with stride 2 to halve the image resolution each time the number of filters is doubled. The resulting 512 feature maps are followed by one sigmoid activation to obtain a probability useful for the classification problem.

\subsection{Training Details}
We trained our det-cGAN with 64 $\times$ 64 resized depth maps as input and simultaneously providing the original RGB images associated with the depth data in order to compute the MSE loss. We adopted the standard approach in \cite{goodfellow2014generative} to optimize the network alternating gradient descent updates between the generator and the discriminator with K = 1. We used mini-batch SGD applying the \textit{Adam} solver with momentum parameters $\beta_1$ = 0.5 and $\beta_2$ = 0.999. In our experiments we chose a $\lambda$ value of $10^{−1}$ in Equation \eqref{eq:c} and a batch size of 64. Some best results are presented in Figure \ref{fig:copertina}.

\begin{figure}[b!]
    \centering
    \includegraphics[width=0.95\columnwidth]{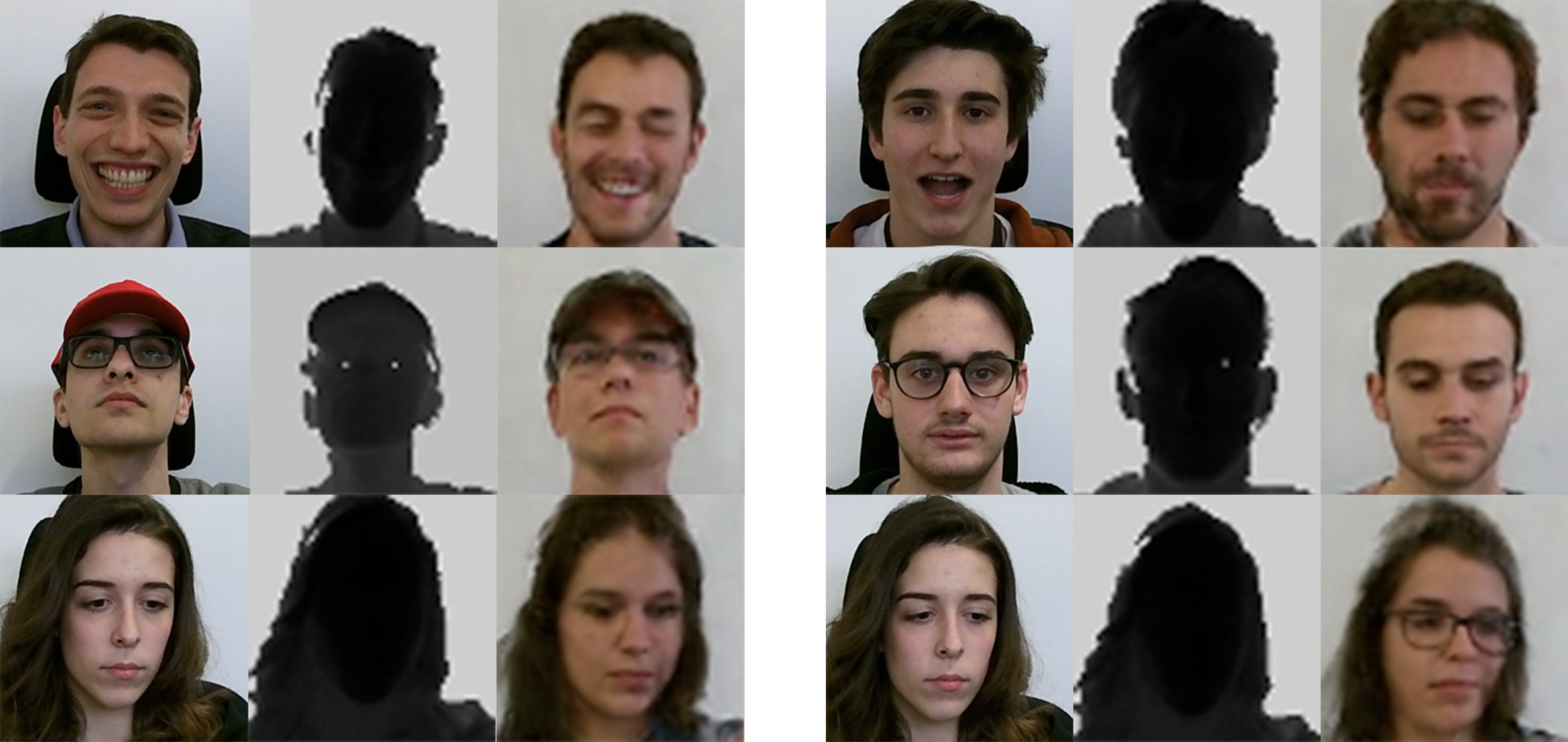}
    \caption{Visual examples of generated images that preserve (left column) and do not preserve (right column) some attributes.}
    \label{fig:attributi}
\end{figure}

\begin{figure}[b!]
    \centering
    \includegraphics[width=0.95\columnwidth]{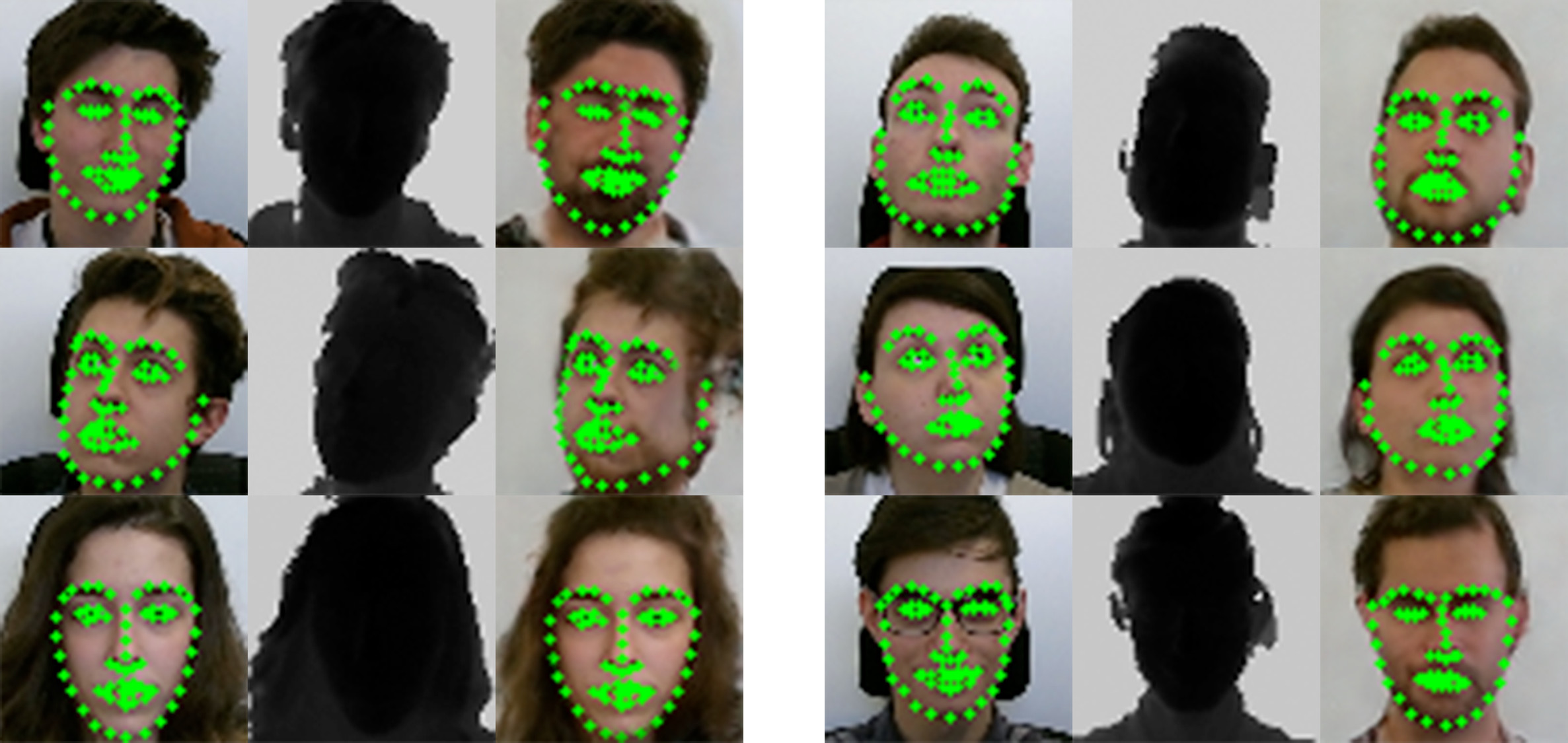}
    \caption{Visual examples of landmark predictions on real and generated images.}
    \label{fig:landmarks}
\end{figure}

\begin{figure}[b!]
    \centering
    \includegraphics[width=0.95\columnwidth]{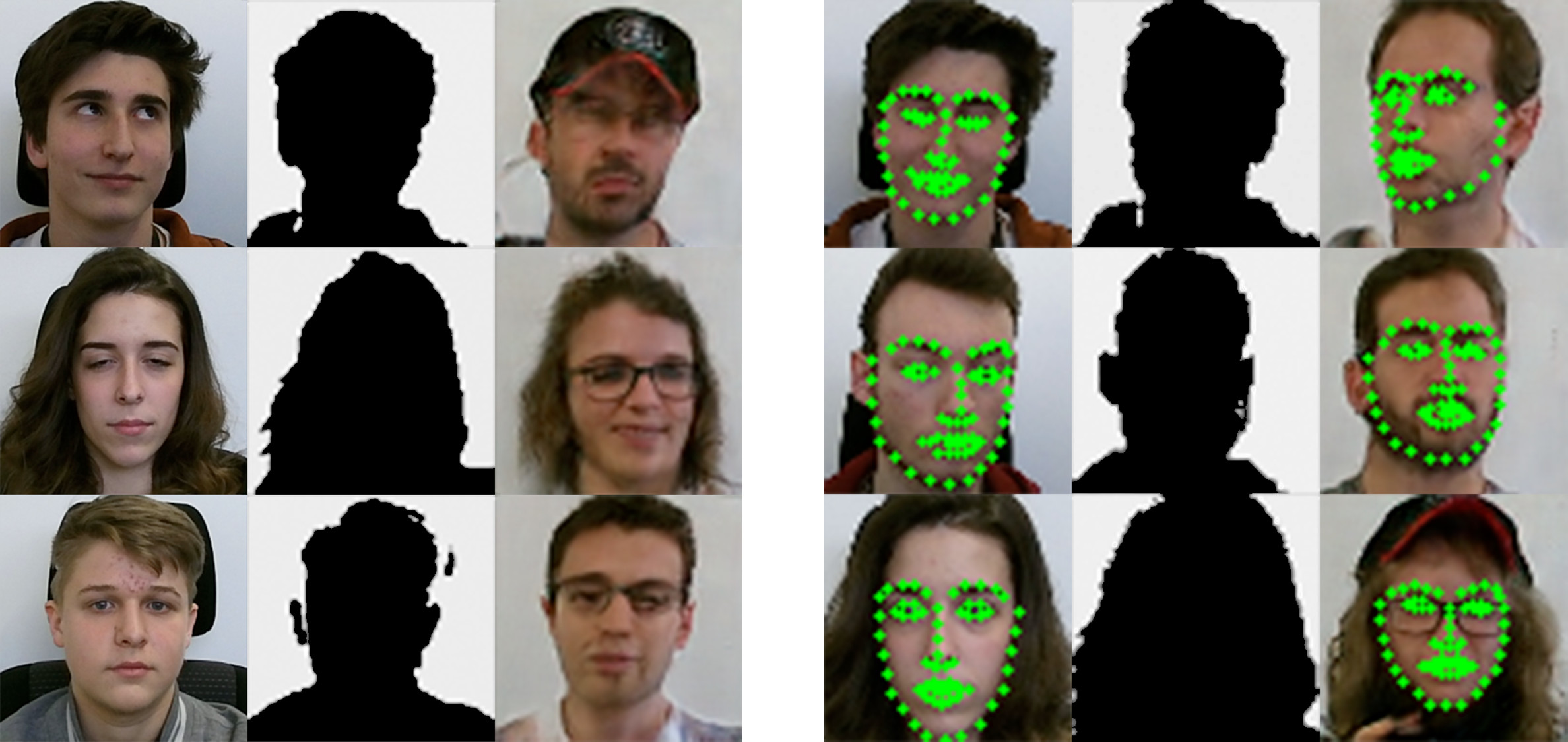}
    \caption{Visual examples of issues using binary maps instead of depth maps: attributes are not preserved (left column) and landmark localization is not precise (right column).}
    \label{fig:binary}
\end{figure}

\begin{table}[t!]
\centering
\caption{Per attribute concordance between the true RGB face and the hallucinated one using VGG-Face CNN.}
\small
\begin{tabular}{c|c|c|c|c}
\hline
\textbf{Attribute}  & \textbf{Accuracy} & \textbf{Precision} & \textbf{Recall} & \textbf{F1} \\
\hline \hline
Male	& 90.30 & 95.51 & 93.49 & 94.49 \\
Young & 93.01 & 97.69 & 95.09 & 96.37  \\
Mouth Open	& 82.86 & 92.16 & 51.07 & 65.71  \\
Smiling	& 96.25 & 99.54 & 66.48 & 79.72  \\
Wearing Hat	& 98.40 & 99.38 & 58.05 & 73.29  \\
Wavy Hair	& 98.46 & 95.28 & 48.44 & 64.24  \\
No Beard	& 48.18 & 63.89 & 40.88 & 49.86  \\
Straight Hair & 79.12 & 07.78 & 57.76 & 13.71  \\
Eyeglasses & 80.12 & 24.91 & 08.14 & 12.27  \\ \hline
\end{tabular}
\label{tab:attributes}
\end{table}

\section{Experiments}
Generally, evaluating the quality of reconstructed images is still an open problem, as reported in \cite{isola2016image}. Traditional metrics such as $L1$ distance are not sufficient to assess joint statistic on the produced images, and therefore do not extrapolate the full structure of the result.
In order to more holistically investigate the capabilities of our network to synthesize RGB face images directly from depth maps, a reconstruction comparison and two perceptual probes are performed.
Firstly, we compared the performance of the proposed model with other \textit{Image-to-Image} recent methods present in the literature, through metrics directly calculated over the reconstructed images.
Secondly, we measured the capability of the proposed network of being able to preserve original facial attributes, like wearing hat and smiling, by exploiting a classification network trained with RGB face images.
Thirdly, we measured whether or not reconstructed RGB face images are realistic enough that an off-the-shelf landmark localization system is able to localize accurate key-points.
Eventually, in order to investigate how much the depth map information impacts the reconstruction task, we repeated the previous experiments testing our network trained with binary maps derived from the original depth maps.

\subsection{Datasets}
Experiments are conducted exploiting two publicly available datasets: \textit{Pandora} \cite{borghi17cvpr} and \textit{MotorMark} \cite{frigieri2017fast}. \textit{Pandora} contains more than 250k frames, splitted into 110 annotated sequences of 22 different actors (10 males and 12 females), while \textit{MotorMark} is composed of more than 30k frames of 35 different subjects, guaranteeing a great variety of face appearances. Subjects can wear garments and sunglasses and may perform driving activities actions like turning the steering wheel, adjust the rear mirror and so on. Both datasets have been acquired with a \textit{Microsoft Kinect One}.
In our experiments, \textit{Pandora} has been used as the training set and \textit{MotorMark} as the test set, performing a cross-dataset validation of the proposed method.\\



\subsection{Reconstruction Comparison}
Here, we check the capabilities of the proposed network to reconstruct RGB images from the correspondent depth ones. We exploited the metrics described in \cite{eigen2014depth}: these metrics were originally used to evaluate depth images generated from RGB image sources.
Results are reported in Table \ref{tab:metrics}. In particular, we compared our generative method with two other techniques: an Autoencoder trained with the same architecture as our Generator network, and \textit{pix2pix} \cite{isola2016image}, a recent work that exploits the Conditional GAN framework. In the last line of Table \ref{tab:metrics}, is also reported the comparison with our network trained on binary maps, detailed at the end of this section. As shown, results confirm the superior accuracy of the presented method. 

\begin{table}[t!]
\centering
\caption{Quantitative comparison about the average attributes concordance between true and hallucinated RGB faces.}
\small
\begin{tabular}{c|c|c|c|c}
\hline
\textbf{Method}  & \textbf{Accuracy} & \textbf{Precision} & \textbf{Recall} & \textbf{F1} \\
\hline \hline
Autoencoder & 75.21 & 61.84 & 40.55 & 51.38  \\ \hline
pix2pix & 84.57 & 74.42 & 56.01 & 60.78\\ \hline
\textbf{Our} & \textbf{85.19} & \textbf{75.13} & \textbf{57.71} & \textbf{61.07}  \\ \hline \hline
Our (Binary Maps) & 60.51 & 49.48 & 29.12 & 42.76  \\ \hline
\end{tabular}
\label{tab:attr2}
\end{table}

\subsection{Attribute Classification}
In the previous section, we checked the overall quality of the reconstructed RGB images. Here, we focus on the capability of our network to generate face images that specifically preserve the facial attributes of the original person.
To this end, we exploited a pre-trained network, the \textit{VGG-Face} CNN \cite{Parkhi15}, trained on RGB images for face recognition purposes. In order to extrapolate only the attributes that can be carried by depth information, we fine-tuned the network with the \textit{CelebA} Dataset \cite{liu2015faceattributes}.

By observing Table \ref{tab:attributes}, it is evident the good capability of the network to preserve gender, age, pose, and appearance attributes. Nevertheless, the depth sensor resolution fails at modeling hair categories such as curly or straight and glasses since such details are not always correctly captured in terms of depth. Glasses lenses, for example, are neglected by IR sensors and significantly captured only when the glasses structure is solid and visible. In all the other cases they tend to be confused by the network with the ocular cavities. Nonetheless, Table \ref{tab:attr2} exhibits the superiority of our proposal against state of the art generative networks also in attribute preservation. Moreover in Figure \ref{fig:attributi} are presented both successful and failure cases.

\subsection{Landmark Localization}
The intuition behind this experiment is that if the synthesized images are realistic and accurate enough, then a landmark localization method trained on real images will be able to localize key-points also on the generated images.
To this aim, we exploited the algorithm included in the \textit{dLib} libraries \cite{dlib09}, which gives landmark positions on RGB images. In Figure \ref{fig:landmarks} qualitative examples that highlight the coherence of landmark predictions between original and generated images are presented. The last column of Table \ref{tab:landmark} reports, for each method, the average $L2$ Norm between the position of landmarks predicted and the ground truth provided by the dataset. The results show that our method is able to produce outputs that can fool an algorithm trained on RGB face images.

\subsection{Binary Maps}
An ablation study is conducted to investigate the importance of depth information, by training our network providing as input binary maps instead of depth maps. 
Binary maps were gathered thresholding the depth maps. Figure \ref{fig:binary} shows examples where the reconstructed face images are not coherent with the original images in terms of attributes preservation and landmark position. At the end of Tables \ref{tab:metrics}, \ref{tab:attr2} and \ref{tab:landmark} are reported the results of the previous experiment where we used binary maps instead of depth maps. Results show that the depth information has a fundamental importance in the face generation task, to preserve coherent facial attributes and head pose orientation.

\begin{table}[t!]
\centering
\caption{Quantitative comparative results of our proposal against the Autoencoder and pix2pix baselines in terms of face detector accuracy and landmark localization.}
\small
\begin{tabular}{c|c|c}
\hline
\textbf{Method}  & \textbf{Accuracy} & \textbf{$L2$ Norm} \\
\hline \hline
Autoencoder	& 54.03 & 2.219 \\ \hline
pix2pix 	& 85.21 & 2.201\\ \hline
\textbf{Our}	& \textbf{86.86} & \textbf{2.089} \\ \hline
 \hline 
Our (Binary Maps) &62.37	& 2.980 \\ \hline
\end{tabular}
\label{tab:landmark}
\end{table}

\section{Conclusion}
In this paper, a deterministic conditional GAN to reconstruct RGB face images from the correspondent depth one is presented. Experimental results confirm the ability of the proposed method to generate accurate faces, to preserve facial attributes and to maintain coherency in facial landmarking. Besides, we checked and shown the importance of depth data in all these tasks.
Various future works can be planned, due to the flexibility and the accuracy of the presented method. For instance, it is possible to investigate how to generate or delete specific face attributes, or how the enhance the training capabilities of depth maps.

\small{
\section*{Acknowledgment}
This work has been carried out within the projects ``COSMOS Prin 2015'' supported by the Italian MIUR, Ministry of Education, University and Research and ``FAR2015 - Monitoring the car drivers attention with multisensory systems, computer vision and machine learning'' funded by the University of Modena and Reggio Emilia. We also acknowledge CINECA for the availability of high-performance computing resources.
}






%



\bibliographystyle{IEEEtran}
\bibliography{biblio}

\begin{thebibliography}{10}
\providecommand{\url}[1]{#1}
\csname url@samestyle\endcsname
\providecommand{\newblock}{\relax}
\providecommand{\bibinfo}[2]{#2}
\providecommand{\BIBentrySTDinterwordspacing}{\spaceskip=0pt\relax}
\providecommand{\BIBentryALTinterwordstretchfactor}{4}
\providecommand{\BIBentryALTinterwordspacing}{\spaceskip=\fontdimen2\font plus
\BIBentryALTinterwordstretchfactor\fontdimen3\font minus
  \fontdimen4\font\relax}
\providecommand{\BIBforeignlanguage}[2]{{%
\expandafter\ifx\csname l@#1\endcsname\relax
\typeout{** WARNING: IEEEtran.bst: No hyphenation pattern has been}%
\typeout{** loaded for the language `#1'. Using the pattern for}%
\typeout{** the default language instead.}%
\else
\language=\csname l@#1\endcsname
\fi
#2}}
\providecommand{\BIBdecl}{\relax}
\BIBdecl

\bibitem{masci2011stacked}
J.~Masci, U.~Meier, D.~Cire{\c{s}}an, and J.~Schmidhuber, ``Stacked
  convolutional auto-encoders for hierarchical feature extraction,''
  \emph{Artificial Neural Networks and Machine Learning--ICANN 2011}, pp.
  52--59, 2011.

\bibitem{mao2016image}
X.-J. Mao, C.~Shen, and Y.-B. Yang, ``Image restoration using convolutional
  auto-encoders with symmetric skip connections,'' \emph{arXiv preprint
  arXiv:1606.08921}, 2016.

\bibitem{bigdeli2017image}
S.~A. Bigdeli and M.~Zwicker, ``Image restoration using autoencoding priors,''
  \emph{arXiv preprint arXiv:1703.09964}, 2017.

\bibitem{guillemot2014image}
C.~Guillemot and O.~Le~Meur, ``Image inpainting: Overview and recent
  advances,'' \emph{IEEE signal processing magazine}, vol.~31, no.~1, 2014.

\bibitem{singhtransforming}
M.~S. Singh, V.~Pondenkandath, B.~Zhou, P.~Lukowicz, and M.~Liwicki,
  ``Transforming sensor data to the image domain for deep learning-an
  application to footstep detection.''

\bibitem{borghi17cvpr}
G.~Borghi, M.~Venturelli, R.~Vezzani, and R.~Cucchiara, ``Poseidon:
  Face-from-depth for driver pose estimation,'' in \emph{Proceedings of the
  IEEE Conference on Computer Vision and Pattern Recognition (CVPR)}, 2017.

\bibitem{goodfellow2014generative}
I.~Goodfellow, J.~Pouget-Abadie, M.~Mirza, B.~Xu, D.~Warde-Farley, S.~Ozair,
  A.~Courville, and Y.~Bengio, ``Generative adversarial nets,'' in
  \emph{Advances in neural information processing systems}, 2014.

\bibitem{radford2015unsupervised}
A.~Radford, L.~Metz, and S.~Chintala, ``Unsupervised representation learning
  with deep convolutional generative adversarial networks,'' \emph{arXiv
  preprint arXiv:1511.06434}, 2015.

\bibitem{isola2016image}
P.~Isola, J.-Y. Zhu, T.~Zhou, and A.~A. Efros, ``Image-to-image translation
  with conditional adversarial networks,'' \emph{arXiv preprint
  arXiv:1611.07004}, 2016.

\bibitem{matteo2017generative}
M.~Fabbri, S.~Calderara, and R.~Cucchiara, ``Generative adversarial models for
  people attribute recognition in surveillance,'' in \emph{14th IEEE
  International Conference on AVSS}, 2017.

\bibitem{zhu2017unpaired}
J.-Y. Zhu, T.~Park, P.~Isola, and A.~A. Efros, ``Unpaired image-to-image
  translation using cycle-consistent adversarial networks,'' \emph{arXiv
  preprint arXiv:1703.10593}, 2017.

\bibitem{borghi2017face}
G.~Borghi, M.~Fabbri, R.~Vezzani, S.~Calderara, and R.~Cucchiara,
  ``Face-from-depth for head pose estimation on depth images,'' \emph{arXiv
  preprint arXiv:1712.05277}, 2017.

\bibitem{frigieri2017fast}
E.~Frigieri, G.~Borghi, R.~Vezzani, and R.~Cucchiara, ``Fast and accurate
  facial landmark localization in depth images for in-car applications,'' in
  \emph{International Conference on Image Analysis and Processing}, 2017.

\bibitem{venturelli2017deep}
M.~Venturelli, G.~Borghi, R.~Vezzani, and R.~Cucchiara, ``Deep head pose
  estimation from depth data for in-car automotive applications,'' \emph{2nd
  International Workshop on Understanding Human Activities through 3D Sensors
  (UHA3DS'16)}, 2016.

\bibitem{lee2012intelligent}
C.-H. Lee, Y.-C. Su, and L.-G. Chen, ``An intelligent depth-based obstacle
  detection system for visually-impaired aid applications,'' in
  \emph{International Workshop on Image Analysis for Multimedia Interactive
  Services}, 2012.

\bibitem{hochreiter1997long}
S.~Hochreiter and J.~Schmidhuber, ``Long short-term memory,'' \emph{Neural
  computation}, vol.~9, no.~8, pp. 1735--1780, 1997.

\bibitem{venturellisiamese}
M.~Venturelli, G.~Borghi, R.~Vezzani, and R.~Cucchiara, ``From depth data to
  head pose estimation: a siamese approach,'' 2017.

\bibitem{saxena2006learning}
A.~Saxena, S.~H. Chung, and A.~Y. Ng, ``Learning depth from single monocular
  images,'' in \emph{Advances in neural information processing systems}, 2006.

\bibitem{saxena2009make3d}
A.~Saxena, M.~Sun, and A.~Y. Ng, ``Make3d: Learning 3d scene structure from a
  single still image,'' \emph{IEEE transactions on pattern analysis and machine
  intelligence}, vol.~31, no.~5, pp. 824--840, 2009.

\bibitem{NIPS2014}
D.~Eigen, C.~Puhrsch, and R.~Fergus, ``Depth map prediction from a single image
  using a multi-scale deep network,'' in \emph{Advances in Neural Information
  Processing Systems 27}.\hskip 1em plus 0.5em minus 0.4em\relax Curran
  Associates, Inc., 2014.

\bibitem{ronneberger2015u}
O.~Ronneberger, P.~Fischer, and T.~Brox, ``U-net: Convolutional networks for
  biomedical image segmentation,'' in \emph{International Conference on Medical
  Image Computing and Computer-Assisted Intervention}.\hskip 1em plus 0.5em
  minus 0.4em\relax Springer, 2015, pp. 234--241.

\bibitem{eigen2014depth}
D.~Eigen, C.~Puhrsch, and R.~Fergus, ``Depth map prediction from a single image
  using a multi-scale deep network,'' in \emph{Advances in neural information
  processing systems}.

\bibitem{Parkhi15}
O.~M. Parkhi, A.~Vedaldi, and A.~Zisserman, ``Deep face recognition,'' in
  \emph{British Machine Vision Conference}, 2015.

\bibitem{liu2015faceattributes}
Z.~Liu, P.~Luo, X.~Wang, and X.~Tang, ``Deep learning face attributes in the
  wild,'' in \emph{Proceedings of International Conference on Computer Vision
  (ICCV)}, 2015.

\bibitem{dlib09}
D.~E. King, ``Dlib-ml: A machine learning toolkit,'' \emph{Journal of Machine
  Learning Research}, vol.~10, pp. 1755--1758, 2009.

\end{thebibliography}

\end{document}